\documentclass[11pt]{article}   
\usepackage{graphicx,colacl}  
\newtheorem{example}{}

\newcommand{\comment}[1]{}

\title{Recognizing Syntactic Errors in the \\ Writing of Second
Language Learners\thanks{This work was supported by NSF Grant \#SRS9416916.}}
\author{David Schneider \and Kathleen McCoy\\Departments of
Linguistics and CIS\\University of Delaware\\Newark, DE 19716\\
\{dschneid, mccoy\}@cis.udel.edu}

\begin{document}

\bibliographystyle{acl}

\maketitle

\abstract{
This paper reports on the recognition component of an intelligent
tutoring system that is designed to help foreign language speakers
learn standard English.  The system models the grammar of the learner,
with this instantiation of the system tailored to signers of American
Sign Language (ASL).  We discuss the theoretical motivations for the
system, various difficulties that have been encountered in the
implementation, as well as the methods we have used to overcome these
problems.  Our method of capturing ungrammaticalities involves
using mal-rules (also called 'error productions').
However, the straightforward addition of some mal-rules causes
significant performance problems with the parser.
For instance, the ASL population
has a strong tendency to drop pronouns and the auxiliary verb `to be'.
Being able to account for these as sentences results in an explosion
in the number of possible parses for each sentence.
This explosion, left unchecked, greatly hampers the performance of the
system.
We discuss how this is handled by taking into account expectations
from the specific population (some of which are captured in our unique
user model).
The different representations of
lexical items at various points in the acquisition process are modeled
by using mal-rules, which obviates the need for multiple lexicons.
The grammar is
evaluated on its ability to correctly diagnose agreement problems in
actual sentences produced by ASL native speakers.  }

\vspace{-.04in}
\section{Overview}
\vspace{-.02in}

This paper reports on the error-recognition component of the ICICLE
(Interactive Computer Identification and Correction of Language
Errors) system.  The system is designed to be a tutorial system for
helping second-language (L2) learners of English.  In this
instantiation of the system, we are focusing on the particular
problems of American Sign Language (ASL) native signers.  The system
recognizes errors by using mal-rules (also called 'error-production rules')
\cite{sleeman82}, \cite{WeiVJ78}
which
extend the language accepted by the grammar to include sentences
containing the specified errors.
The mal-rules themselves are derived from an error taxonomy which was
the result of an analysis of writing samples.
This paper focuses primarily  on the
unique challenges posed by developing a grammar that allows the parser
to efficiently parse and recognize errors in sentences even when multiple 
errors occur.
Additionally, it is important to note that the users will not be at a
uniform stage of acquisition $-$ the system must be capable of processing
the input of users with varying levels of English competence.  We briefly describe how acquisition is modeled and how this model can help
with some of the problems faced by a system designed to recognize
errors. 
\vspace{-.04in}
\vspace{-.04in}

We will begin with an overview of the entire ICICLE system.  To
motivate some of the difficulties encountered by our mal-rule-based
error recognition system, we will briefly describe some of the errors
common to the population under study.  A major problem that must be
faced is parsing efficiency caused by multiple parses.  This is a
particularly difficult problem when expected errors include omission
errors, and thus this class of errors will be discussed in some
detail.  Another important problem involves the addition/subtraction
of various syntactic features in the grammar and lexicon during
acquisition.  We describe how our system models this without the use
of multiple lexicons.  We follow this by a description of the current
implementation and grammar coverage of the system.  Finally, we will
present an evaluation of the system for number/agreement errors in the
target group of language learners.

\vspace{-.04in}
\section{System Overview}
\vspace{-.02in}

The ICICLE system is meant to help second-language learners by
identifying errors and engaging the learners in a tutorial dialogue.
It takes as input a text written by the student.
This is given to the error identification component, which is
responsible for flagging the errors.
The identification  is done by parsing the input one sentence at a
time using a bottom-up chart parser which is a successor to \cite{allen95}.
The grammar formalism used by the parser consists of context-free rules
augmented with 
features.
The grammar itself is a grammar of English which has been augmented
with a set of mal-rules which capture errors common to this
user population.
We will briefly discuss some classes of errors that were uncovered in
our writing sample analysis which was used to
identify errors expected in this population.
This discussion will motivate some of the 
mal-rules which were written to capture some classes of errors, and the
difficulties encountered in implementing these mal-rules.
The mal-rules are specially tagged with information helpful in the
correction phase of the system.

The error identification component relies on information in the user
model $-$ the most interesting aspect of which is a model of the acquisition
of a second language. This model (instantiated with information from the ASL/English
language model) is used to highlight those grammar rules which the
student has most likely already acquired or is currently in the
process of acquiring. These rules will be the ones the parser attempts to use when
parsing the user's input. Thus we take an interlanguage view of the acquisition process
\cite{selinker72}, \cite{ellis94}, \cite{cook93} and attempt to model
how the student's grammar is likely to change over time.  The essence
of the acquisition model is that there are discrete stages that all
learners of a particular language will go through \cite{krashen81},
\cite{ingram89}, \cite{DB74},
\cite{BMK74}.  Each of these stages is characterized
in our model by sets of language features (and therefore
constructions) that the learner is in the process of acquiring.  It is
anticipated that most of the errors that learners make will be
within the constructions (where ``construction'' is construed broadly) that they are in the process of acquiring
\cite{vygotsky86} and that they will favor sentences involving
those constructions in a ``hypothesize and test'' style of learning, as
predicted by interlanguage theory.  Thus, the parser favors grammar
rules involving constructions currently being acquired (and, to a lesser
extent, constructions already acquired).

The correction phase of the system is a focus of current research.
A description of the strategies for this phase can be
found in 
\cite{Michaud&McCoy98} and \cite{Michaud98}.


\vspace{-.04in}
\section{Expected Errors}
\vspace{-.02in}
In order to identify the errors we expect the population to make, we collected writing samples from a number of different schools
and organizations for the deaf.  To help
identify any instances of language transfer between ASL and written
English, we concentrated on eliciting samples
from deaf people who are native ASL signers.  It is important to note that ASL is not simply a translation
of standard English into manual gestures, but rather is a complete
language with its own syntax, which is significantly different from
English.  
Some of our
previous work  \cite{journSuriMcCoy93}
explored how language transfer might influence written English and
suggested that negative language transfer might occur when the
realization of specific language features differed between the first
language and written English.
For instance, one feature is the realization of the copula ``be''.  In
ASL the copula ``be'' is often not lexicalized.  Thus, negative
language transfer might predict omission errors resulting from not
lexicalizing the copula ``be'' in the written English of ASL signers.
While we concentrate here on errors from the ASL population,
the errors identified
are likely to be found in learners coming from first languages other
than ASL as well.  This would be the case if the first language has
features in common with
ASL.  For instance the missing copula ``be'' is also a common error in
the writing of native Chinese speakers since Chinese and ASL share the
feature that the copula ``be'' is often not lexicalized.
Thus, the examples seen here will generalize to other languages.

In the following we describe some classes of errors which we uncovered
(and 
attempt to ``explain'' why an ASL native might come to make these
errors). 

\vspace{-.04in}
\subsection{Constituent Omissions}
\vspace{-.04in}
\label{asl-omissions}
Learners of English as a second language (ESL) omit constituents for a variety of reasons.  One error that is common for
many ASL learners is the dropping of determiners.
Perhaps because ASL does not have a determiner system similar to that
of English, it is not unusual for a determiner to be omitted as
in: 
\vspace{-.1in}
\begin{example}
{
I am \verb+_+ transfer student from ....
}
\end{example}
\vspace{-.1in}

These errors can be flagged reasonably well when they
are syntactic (and not pragmatic) in nature and do not pose  much
additional burden on the parser/grammar.

However, missing main verbs (most commonly missing copulas) are also common in our writing samples:
\vspace{-.1in}
\begin{example}
{
Once the situation changes they \verb+_+ different people.
}
\end{example}
\vspace{-.1in}

One explanation for this (as well as other missing elements such as
missing prepositions) is that copulas are not overtly lexicalized in
ASL because the copula (preposition) is gotten across in different
ways in ASL.
Because the copula (preposition) is realized in a radically different
fashion in ASL, there can be no positive language
transfer for these constructions.

In addition to omitting verbs, some NPs may also be omitted.  It has
been argued (see, for example \cite{LM91}) that ASL allows
topic NP deletion \cite{Huang84} which means that topic noun phrases
that are prominent in the discourse context may be left out of a
sentence.  Carrying this strategy over to English might explain why
some NPs are omitted from sentences such as:

\vspace{-.1in}
\begin{example}
{
While living at college I spend lot of money because \verb+_+ go out to eat
almost everyday.
}
\end{example}
\vspace{-.1in}

Mal-rules written to handle these errors must capture missing verbs, NPs, and
prepositions.  The grammar is further complicated because ASL natives
also have many errors in relative clause formation including missing
relative pronouns.  The possibility of all of these omissions causes
the parser to explore a great number of parses (many of which will
complete successfully). 

\vspace{-.04in}
\subsection{Handling Omissions}
\vspace{-.02in}
As we just saw, omissions are frequent in the writing of ASL
natives and they are difficult to detect using the mal-rule formalism.  To
clearly see the problem, consider the following two sentences, which
would not be unusual in the writing of an ASL native.

\vspace{-.1in}
\begin{example}
{
\label{boyhappy}
The boy happy.
}
\end{example}
\vspace{-.1in}
\vspace{-.1in}
\begin{example}
{
\label{ishappy}
Is happy.
}
\end{example}
\vspace{-.1in}

As the reader can see, in \ref{boyhappy} the main verb ``be'' is
omitted, while the subject is missing in \ref{ishappy}.

To handle these types of
sentences, we included in our grammar mal-rules like the following:

\vspace{-.1in}
\begin{example}
\label{vpadjp}
{\small
{\tt VP(error +)  $\rightarrow$ AdjP}
}
\end{example}
\vspace{-.1in}
\vspace{-.1in}
\begin{example}
{\small
\label{SVP}
{\tt S(error +)  $\rightarrow$ VP}
}
\end{example}
\vspace{-.1in}

A significant problem that arises from these rules is that a simple
adjective is parsed as an S even if it is in a normal, grammatical
sentence.  This behavior leads to many extra parses, since the S will
be able to participate in lots of other parses.  The problem becomes
much more serious when the other possible omissions are added into the
grammar.  However, closer examination of our writing samples indicates
that, except for determiners, our users generally leave
out at most one word (constituent) per sentence.  Thus it is unlikely
that ``happy'' will ever be an entire sentence.  We would like this
fact to be reflected in the analyses explored by the parser.  However,
a traditional bottom-up context-free parser has no way to deal with
this case, as there is no way to block rules from firing as long as
the features are capable of unification.

One possibility would be to allow the {\tt (error +)} feature to
percolate up through the parse.  Any rule which introduces the {\tt
(error +)} feature could then be prevented from having any children
specified with {\tt (error +)}.  However, this solution would be far
too restrictive, as it would restrict the number of errors in a
sentence to one, and many of the sentences in our ASL corpus involve
multiple errors.

Recall, however, that in our analysis we found that (except for
determiners) our writing samples did not contain multiple omission
errors in a sentence.  Thus another possibility might be to percolate
an error feature associated with omissions only$-$perhaps called {\tt
(missing +)}.

Upon closer inspection, this solution also has difficulties.  The
first difficulty has to do with implementing the feature percolation.
For instance, for a VP to be specified as {\tt(missing +)} whenever any of its
sub-constituents has that feature, one would need to have separate rules
raising the feature up from each of the sub-constituents, as in the
following:

\vspace{-.1in}
\begin{example}
{\small
\label{vpvnpnpgap}
{\tt VP(missing ?a) $\rightarrow$ V NP NP(missing ?a)}
}
\end{example}
\vspace{-.1in}
\vspace{-.1in}
\begin{example}
{\small
\label{VpVNpGapNp}
{\tt VP(missing ?a) $\rightarrow$ V NP(missing ?a) NP}
}
\end{example}
\vspace{-.1in}
\vspace{-.1in}
\begin{example}
{\small
\label{VpVGapNpNp}
{\tt VP(missing ?a) $\rightarrow$ V(missing ?a) NP NP}
}
\end{example}
\vspace{-.1in}

This would cause an unwarranted increase in the size of the grammar,
and would also cause an immense increase in the number of parses,
since three VPs would be added to the chart, one for each of the rules.  

At first glance it appears that this problem can be overcome with the
use of ``foot features,'' which are included in the parser we are
using.  A foot feature moves features from any child to the parent.
For example, for a foot feature F, if one child has a specification
for F, it will be passed on to the parent.  If more than one child is
specified for F, then the values of F must unify, and the unified
value will be passed up the parent.  While the use of foot features
appears to make the feature percolation easier, it will not allow the
feature to be used as desired.  In particular, we need to have the
feature percolated {\em only} when it has a positive value and only
when that value is associated with exactly one constituent on the
right-hand side of a rule.  The foot feature as defined by the parser
would allow the percolation of the feature even if it were specified
in more than one constituent.

A further complication with using this type of feature
propagation arises because there are some situations where multiple
omission errors do occur, especially when determiners are
omitted.\footnote{\label{determiner_omit}While our analysis so far has only indicated that
determiner omissions have this property, we do not want to rule out the
possibility that other combinations of omission errors might be found
to occur as well.}  Consider the following example taken from our
corpus where both the main verb ``be'' and a determiner ``the'' are omitted.

\vspace{-.1in}
\begin{example}
{
\label{mult-omission}
Student always bothering me while I am at dorm.\\
(Corrected) Student\underline{s} \underline{are} always bothering me while I am at \underline{the}
dorm.
}
\end{example}
\vspace{-.1in}

Our solution to the problem involves using procedural attachment.  The
parser we are using builds constituents and stores them in a chart.
Before storing them in the chart, the parser can run
arbitrary procedures on new constituents.  These procedures, specified
in the grammar, will be run on all
constituents that meet a certain pattern specified by the grammar
writer.

Our procedure amounts to specifying an alternative method for
propagating the {\tt (missing +)} feature, which will still be a foot feature.  It will be run on any
constituent that specifies {\tt (missing +)}.  The procedure can either delete a
constituent that has more than one child with {\tt (missing +)}, or it
can alter the {\tt (missing +)} feature on the constituent in the
face of determiner omissions (as discussed in footnote \ref{determiner_omit}).
By using a special procedure to implement the feature percolation, we
will be able to be more flexible in where we allow the ``missing''
feature to percolate.

\vspace{-.04in}
\subsection{Syntactic Feature Addition}
\vspace{-.02in}

For this system to properly model language acquisition, it must also
model the addition (and possible subtraction) of syntactic features in
the lexicon and grammar of the learner.  For instance, ASL natives
have a great deal of difficulty with many of the agreement features in
English.  As a concrete 
example, this population frequently has trouble with the difference
between ``other'' and ``another''.  They frequently use ``other'' in a singular NP, where ``another'' would normally be
called for.  We hypothesize that this is partly a result of their not
understanding that there is agreement between NPs and their
specifiers (determiners, quantifiers, etc.).  Even if this is
recognized, the learners may not
have the lexical representations necessary to support the agreement
for these two words.\footnote{``Another'' and ``other'' are not
separate lexical items in ASL.}
 Thus, the most accurate model of
the language of these early learners involves a lexicon with
impoverished entries $-$ i.e. no person or number features for
determiners and quantifiers.  Such an impoverished lexicon would mean
that the entries for the two words might be identical, which appears
to be the case for these learners.  

There are at least two reasons for not using this sort of
impoverished lexicon.  Firstly, it would require having multiple
lexicons (some impoverished, others not), with the system needing to
determine which to use for a given user.  Secondly, it would not allow
grammatical uses of the impoverished items to be differentiated from
ungrammatical uses.  With an impoverished lexicon, any use
(grammatical or not) of ``other'' or ``another'' would be flagged as
an error, since it would involve using a lexical entry that does not
have all of the features that the standard entry has.  Since the
lexical item would not have the {\tt agr} specification, it could
not match the rule that requires agreement between determiners and nouns.

\subsubsection{Implementation}
For these reasons, we decided not to use different lexical entries to
model the different stages of acquisition.  Instead, we use
mal-rules, the same mechanism that we are using to model syntactic
changes.  A standard (grammatical) DP (Determiner Phrase) rule has the
following format:

\vspace{-.1in}
\begin{example}
{\small
\label{DPDetNP}
{\tt DP(agr ?a) $\rightarrow$ Det(agr ?a) NP(agr ?a)}
}
\end{example}
\vspace{-.1in}

We initially tried simply eliminating the references to agreement
between the NP and the determiner, as in the following mal-rule:

\vspace{-.1in}
\begin{example}
{\small
\label{Mal-DPDetNP}
{\tt DP(error +)(agr ?a) $\rightarrow$ Det  NP(agr ?a)}
}
\end{example}
\vspace{-.1in}

This has the advantage of flagging any deviant DPs as having the error
feature, since ungrammatical DPs will trigger the mal-rule \ref{Mal-DPDetNP},
but won't trigger \ref{DPDetNP}.
However, a grammatical DP (e.g. ``another child'') fires both the
mal-rule \ref{Mal-DPDetNP} and the grammatical rule \ref{DPDetNP}.
Not only did this behavior cause the parser to slow down very
significantly, since it effectively doubled the number of DPs in a
sentence, but it also has the potential to report an error when one
does not exist.  We also briefly considered using impoverishment rules
on specific categories.  For example, we could have used a rule
stating that determiners have all possible agreement values.  This has
the effect of eliminating agreement as a barrier to unification, much
as would be expected if the learner has no knowledge of agreement on
determiners.  However, this solution has a problem very similar to
that of the previous possible solution: all determiners in the input
could suddenly have two entries in the chart -- one with the actual
agreement, one with the impoverished agreement.  These would then both
be used in parsing, leading to another explosion in the number of
parses.

We finally ended up building a set of rules that matches just the
ungrammatical possibilities, i.e. they do not allow a grammatical
structure to fire both the mal-rule and the normal rule.  The present
set of rules for determiner-NP agreement include the following:

\vspace{-.1in}
\begin{example}
{\small
{\tt DP(agr ?a) $\rightarrow$ Det (agr ?a) NP (agr ?a)}
}
\end{example}
\vspace{-.1in}
\vspace{-.1in}
\begin{example}
{\small
{\tt DP(agr s)(error +) $\rightarrow$ Det(agr (?!a s)) 
 NP(agr s)}
}
\end{example}
\vspace{-.1in}
\vspace{-.1in}
\begin{example}
{\small
{\tt DP(agr p)(error +) $\rightarrow$ Det(agr (?!a p)) NP(agr p)}
}
\end{example}
\vspace{-.1in}
	
This solution required using the negation operator ``!'' present in our
parser to specify that a {\tt Det} not allow singular/plural
agreement.  However, this feature is limited in the present
implementation to constant values, i.e. we can't negate a variable.  This solution achieves the major goal of
not introducing extraneous parses for grammatical constituents.
However, it achieves this goal at some cost.  Namely, we are forced to
increase
the number of rules in order to accomplish the task.  

\subsubsection{Future plans}
\label{feature-addition-future-plans}
We are presently working on the implementation of a variant of unification that will
allow us to do the job with fewer rules.  The new operation will work in the following
sort of rule:

\vspace{-.1in}
\begin{example}
{\small
{\tt DP (agr ?a)$ \rightarrow $ Det(agr ?!a) NP(agr ?a)}
}
\end{example}
\vspace{-.1in}

This rule will be interpreted as follows: the {\tt agr} values between
the \texttt{DP} and the \texttt{NP} will be the same, and none of the
values in \texttt{Det} will be allowed to be in the agreement values
for the \texttt{NP} and the \texttt{DP}.  This will allow the rule to
fire precisely when there are no possible ways to unify the values
between the \texttt{Det} and the \texttt{NP}, i.e. none of the
\texttt{agr} values for the \texttt{Det} will be allowed in the
variable \texttt{?a}.  Thus, this rule will
only fire for ungrammatical constructions.

\vspace{-.04in}
\section{Grammar Coverage/User Interface}
\vspace{-.02in}
\label{coverage}
The ICICLE grammar is a broad-coverage grammar designed to parse
a wide variety of both grammatical sentences and sentences containing
errors.  It is built around the COMLEX
Syntax 2.2 lexicon \cite{grishman-et-al94}, which contains
approximately 38,000 different syntactic head words.  We have a simple
set of rules that allows for inflection, thereby doubling the number
of noun forms, while giving us three to four times as many verb forms
as there are heads.  Thus we can handle approximately 40,000 noun
forms, 8,000 adjectives, and well over 15,000 verb forms.  In addition,
unknown words coming into the system are assumed to be proper nouns,
thus expanding the number of words handled even further.

The grammar itself contains approximately 25 different adjectival
subcategorizations, including subcategorizations requiring an
extraposed structure (the ``it'' in ``it is true that he is here'').
We also include half a dozen noun complementation types.
We have approximately 110 different verb complementation frames, many
of which are indexed for several different subcategorizations.  The grammar is also
able to account for verb-particle constructions when the verb is
adjacent to the particle, as well as when they are separated (e.g. ``I
called him up'').  

Additionally, the grammar allows for various different types of
subjects, including infinitivals with and without subjects (``to fail
a class is unfortunate'', ``for him to fail the class is irresponsible'').
It handles yes/no questions, wh-questions, and both subject and object
relative clauses.

The grammar has only limited abilities concerning coordination $-$ it
only allows limited constituent coordination, and does not allow
non-constituent coordination (e.g. ``I saw and he hit the ball'')
at all.  It is also fairly weak in its handling of adjunct subordinate
clauses.  The population we are concerned with also has significant
trouble with this, in particular there is a strong propensity towards
over-using ``because''.   Adverbs are also problematic, in that the
system is not yet able to differentiate what position a given adverb
should be able to take in a sentence, thus no errors in adverb
placement can be flagged.  We are presently in the process of integrating a new
version of the lexicon that includes features specifying what each adverb
can attach to.  Once this is done, we expect to be able to process
adverbs quite effectively.

The user interface presently consists of a main window where the user can input
the text and control parsing, file access, etc.  After parsing, the
sentences are highlighted with different colors corresponding to
different types of errors.  When the user double-clicks on a sentence,
a separate ``fix-it'' window is displayed with the sentence in
question, along with descriptions of the errors.  The
user can click on the errors and the system will highlight the part
of the sentence where the error occurred.  For example, in the
sentence ``I see a boys'', only ``a boys'' will be
highlighted.  The ``fix-it'' window also allows the user to change the
sentence and then re-parse it.  If the changes are acceptable to the user, the
new sentence can be substituted back into the main text.

\vspace{-.04in}
\section{Evaluation of Error Recognition}
\vspace{-.02in}
An evaluation of the grammar was conducted on a variety of sentences
pulled from the corpus of ASL natives.  The corpus contains essays
written by ASL natives which is annotated with references to different
types of errors in the sentences. The focus for this paper was on
recognition of agreement-type problems, and as such we pulled out all of the sentences
that had been marked with the following errors:
\begin{itemize}
\vspace{-.1in}
\item{NUM: Number problems, which are typically errors in subject-verb
agreement}
\vspace{-.02in}
\item{ED: extra determiner}
\vspace{-.02in}
\item{MD: missing determiner for an NP that requires a determiner}
\vspace{-.02in}
\item{ID: incorrect determiner}
\vspace{-.1in}
\end{itemize}

In addition to testing sentences with these problems, we also tested
fully grammatical sentences from the same corpus, to see if we could
correctly differentiate between grammatical and ungrammatical
sentences that might be produced by our target user group.

After gathering the sentences from the database, we cut them down to
mono-clausal sentences wherever possible, due to the fact that the
handling of adjunct clauses is not yet complete (see
\S\ref{coverage}).  An example of the type of sentence that had to be
divided is the following:

\vspace{-.1in}
\begin{example}
{
They should communicate each other because the communication is very
    important to understand each other.
}
\end{example}
\vspace{-.1in}

This sentence was divided into ``They should communicate each other''
and ``the communication is very important to understand each other.''
In addition to separating the clauses, we also fixed the spelling
errors in the sentences to be tested since spelling correction is
beyond the scope of the current implementation.
\vspace{-.04in}

\subsection{Results for Ungrammatical Sentences}
\vspace{-.04in}

We ended up with 79 sentences to test for the determiner and agreement
errors.  Of these 79 sentences, 44 (56\%) parse with the expected type of
error.  Another 23 (29\%) have no parses that cover the entire sentence, and
12 (15\%) parse as having no errors at
all. 

A number of the sentences that had been flagged with errors in the
database were actually grammatical sentences, but were deemed
inappropriate in context.  Thus, sentences like the following were
tagged with errors in the corpus:

\vspace{-.1in}
\begin{example}
{
I started to attend the class last Saturday.
}
\end{example}
\vspace{-.1in}

It was evident from the context that this sentence should have had ``classes'' rather than ``the class.''
Of the 12 sentences that were parsed as error-free, five were
actually syntactically and semantically acceptable, but
were inappropriate for their contexts, as in the previous example.  Another four had
pragmatic/semantic problems, but were syntactically well-formed, as in

\vspace{-.1in}
\begin{example}
{
I want to succeed in jobs anywhere.
}
\end{example}
\vspace{-.1in}

Thus, there are really only three sentences that do not
have a parse with the appropriate error.  Since this parser is a
syntactic parser, it should not be expected to find the
semantic/pragmatic errors, nor should it know if the sentence was
inappropriate for its context in the essay.  If we
eliminate the nine sentences that are actually grammatical in isolation, we are left with 70
sentences, of which 44 (63\%) have parses with the expected error,
three (4\%) are wrongly accepted as grammatical, and 23 (33\%) do not parse.

In terms of evaluating these results for the purposes of the system,
we must consider the implications of the various categories.  63\%
would trigger tutoring, and 33\% would be tagged as problematic, but would
have no information about the type of error.  In only 4\% of sentences
containing errors would the system incorrectly indicate that no errors
are present.

\vspace{-.04in}
\subsection{Results for Grammatical Sentences}
\vspace{-.04in}

We also tested the system on 101 grammatical sentences that were
pulled from the same corpus.  These sentences were modified in the
same way as the ungrammatical ones, with multi-clausal sentences being
divided up into mono-clausal sentences.  Of these 101 sentences, 89 (88\%)
parsed as having no errors, 3 (3\%) parsed with errors, and the remaining
8 (8\%) did not parse.

The present implementation of the grammar suffers from poor
recognition of coordination, even within single clauses.  Five of the
eleven sentences that did not return an error-free parse suffered
from this limitation.  We expect to be able to improve the numbers
significantly by including in the grammar some recognition of
punctuation, which, due to technical problems, is presently filtered
out of the input before the parser has a chance to use it.

\vspace{-.04in}
\section{Conclusions and Future Work}
\vspace{-.02in}

Future work will include extending the grammar to better deal with
coordination and adjunct clauses.  We will also continue to work on
the negation operator  and the propagation of the {\tt
missing} feature discussed above.  In order to cut down on the number
of parses, as well as to make it easier to decide which is the
appropriate parse to correct, we have recently switched to a
best-first parsing strategy.  This should allow us to model which
rules are most likely to be used by a given user, with the mal-rules corresponding to
the constructions currently being acquired having a higher probability than
those that the learner has already mastered.  However, at the
moment we have simply lowered the probabilities of all mal-rules, so
that any grammatical parses are generated first, followed by the
``ungrammatical'' parses.

As we have shown, this system does a good job of flagging
ungrammatical sentences produced by the target population, with a high
proportion of the flagged sentences containing significant information
about the type and location of the error.
Our continuing work will hopefully improve these percentages, and
couple this recognition component with an intelligent tutoring phase.

\bibliography{/usa/mccoy/papers/treffile,/usa/suri/Bibliography/big}
\end{document}